\documentclass[conference,a4paper]{IEEEtran}
\IEEEoverridecommandlockouts

\ifCLASSINFOpdf
\else
\fi

\usepackage[dvips]{graphicx}
\usepackage{amsmath,amssymb}
\usepackage{paralist}
\usepackage{subfigure}
\usepackage{url}
\usepackage{algorithm}
\usepackage{algorithmic}
\usepackage{fancyhdr}

\makeatletter
\def\tbcaption{\def\@captype{table}\caption}
\def\figcaption{\def\@captype{figure}\caption}

\newcommand{\bvec}[1]{\mbox{\boldmath $#1$}}

\begin{document}

\title{Shortening Time Required for Adaptive Structural Learning Method of Deep Belief Network\\with Multi-Modal Data Arrangement
\thanks{\copyright 2017 IEEE. Personal use of this material is permitted. Permission from IEEE must be obtained for all other uses, in any current or future media, including reprinting/republishing this material for advertising or promotional purposes, creating new collective works, for resale or redistribution to servers or lists, or reuse of any copyrighted component of this work in other works.}
}

\author{
\IEEEauthorblockN{Shin Kamada}
\IEEEauthorblockA{Graduate School of Information Sciences, \\
Hiroshima City University\\
3-4-1, Ozuka-Higashi, Asa-Minami-ku,\\
Hiroshima, 731-3194, Japan\\
Email: da65002@e.hiroshima-cu.ac.jp}
\and
\IEEEauthorblockN{Takumi Ichimura}
\IEEEauthorblockA{Faculty of Management and Information Systems,\\
Prefectural University of Hiroshima\\
1-1-71, Ujina-Higashi, Minami-ku,\\
Hiroshima, 734-8558, Japan\\
Email: ichimura@pu-hiroshima.ac.jp}
}

\maketitle

\begin{abstract}
Recently, Deep Learning has been applied in the techniques of artificial intelligence. Especially, Deep Learning performed good results in the field of image recognition. Most new Deep Learning architectures are naturally developed in image recognition. For this reason, not only the numerical data and text data but also the time-series data are transformed to the image data format. Multi-modal data consists of two or more kinds of data such as picture and text. The arrangement in a general method is formed in the squared array with no specific aim. In this paper, the data arrangement are modified according to the similarity of input-output pattern in Adaptive Structural Learning method of Deep Belief Network. The similarity of output signals of hidden neurons is made by the order rearrangement of hidden neurons. The experimental results for the data rearrangement in squared array showed the shortening time required for DBN learning.\end{abstract}


\IEEEpeerreviewmaketitle

\section{Introduction}
\label{sec:introduction}
Recently, Deep Learning has been applied in the techniques of artificial intelligence \cite{Bengio09,webmarket2016}. Especially, Deep Learning performed good results in the field of image recognition \cite{AlexNet, GoogleNet, Kamada16_SMC, Kamada16_ICONIP, Kamada16_TENCON, Ichimura17_IJCNN}. Most new Deep Learning architectures are naturally developed in image recognition. For this reason, not only numerical data, text data, and binary data but also the time-series data are transformed to the image data format which is trained by Deep Learning.

The various data are collected from all over the world nowadays, because the Internet of Things (IoT) refers to devices that are connected to the Internet and can send and receive data. Their collected data can be seen as a kind of multi modal data. Multi-modal data consists of two or more kinds of data such as picture and text. For example, a twitter user can post a picture and the message at once. The posted data are displayed on the time line. The collection of such data and its analysis discovers a new knowledge and information to deliver to you \cite{Ichimura_IJKWI}.

However, the data arrangement in a general Deep Learning method is formed in the squared array with no specific aim. In other words, we may meet a problem of the combination of two or more kind of data in data arrangement, because we have to compose a new arrangement which represents the relation among two or more data having the feature of the respective original data.

A framework for encoding time series into different types of images has been proposed \cite{Wang_IJCAI15}. The method is Gramian Angular Summation/ Difference Fields (GASF/GADF) and Markov Transition Fields (MTF) to classify time series images by polar coordinate.

Topological data analysis (TDA) is an approach to the analysis of datasets using techniques from topology. TDA can extract robust topological features from data and use these summaries for modeling the data. Fujitsu Co.Ltd. develops a representing method of a feature of time series data by using the chaos theory and TDA \cite{Fujitsu}.

However, their frameworks should be required to analyze the data before training on Deep Learning, and then the analysis takes long time and the reconstruction of an image is not executable. We consider a data arrangement method which is modified according to the similarity of input-output pattern in Adaptive Structural Learning method of Deep Belief Network (DBN) \cite{Kamada16_SMC, Kamada16_ICONIP, Kamada16_TENCON, Ichimura17_IJCNN}. DBN \cite{Hinton06} can be viewed as a composition of simple, unsupervised networks such as Restricted Boltzmann Machines (RBMs) \cite{Hinton12}. The adaptive learning method can generate a new neuron of RBM, if the classification power is insufficient in the RBM structure. The similarity of output signals of hidden neurons is made by the order rearrangement of hidden neurons. The experimental results for the data rearrangement in squared array showed the shortening time required for DBN learning.

\section{Adaptive Learning Method of DBN}
\label{sec:adaptive_dbn}
RBM \cite{Hinton12} is a stochastic unsupervised learning model. The network structure of RBM consists of two kinds of binary layers: one is a visible layer $\bvec{v} \in \{0, 1 \}^{I}$ with the parameter $\bvec{b} \in \mathbb{R}^{I}$ for input patterns, and the other is a hidden layer $\bvec{h} \in \{0, 1 \}^{J}$ with the parameter $\bvec{c} \in \mathbb{R}^{J}$ to represent the feature of input space. $I$ and $J$ are the number of visible and hidden neurons, respectively. The connection between a visible neuron $v_{i}$ and a hidden neuron $h_j$ is represented as the weight $W_{ij}$. There are not any connections among the neurons in the same layer. The objective of RBM learning is to minimize the energy function $E(\bvec{v}, \bvec{h})$ that is defined as follows, with the learning parameters $\bvec{\theta}=\{\bvec{b}, \bvec{c}, \bvec{W} \}$. 
\begin{equation}
E(\bvec{v}, \bvec{h}) = - \sum_{i} b_i v_i - \sum_j c_j h_j - \sum_{i} \sum_{j} v_i W_{ij} h_j ,
\label{eq:energy}
\end{equation}
\begin{equation}
p(\bvec{v}, \bvec{h})=\frac{1}{Z} \exp(-E(\bvec{v}, \bvec{h})), \;  Z = \sum_{\bvec{v}} \sum_{\bvec{h}} \exp(-E(\bvec{v}, \bvec{h})) ,
\label{eq:prob}
\end{equation}

\begin{figure}[htbp]
\begin{center}
\subfigure[Neuron Generation]{\includegraphics[scale=0.5]{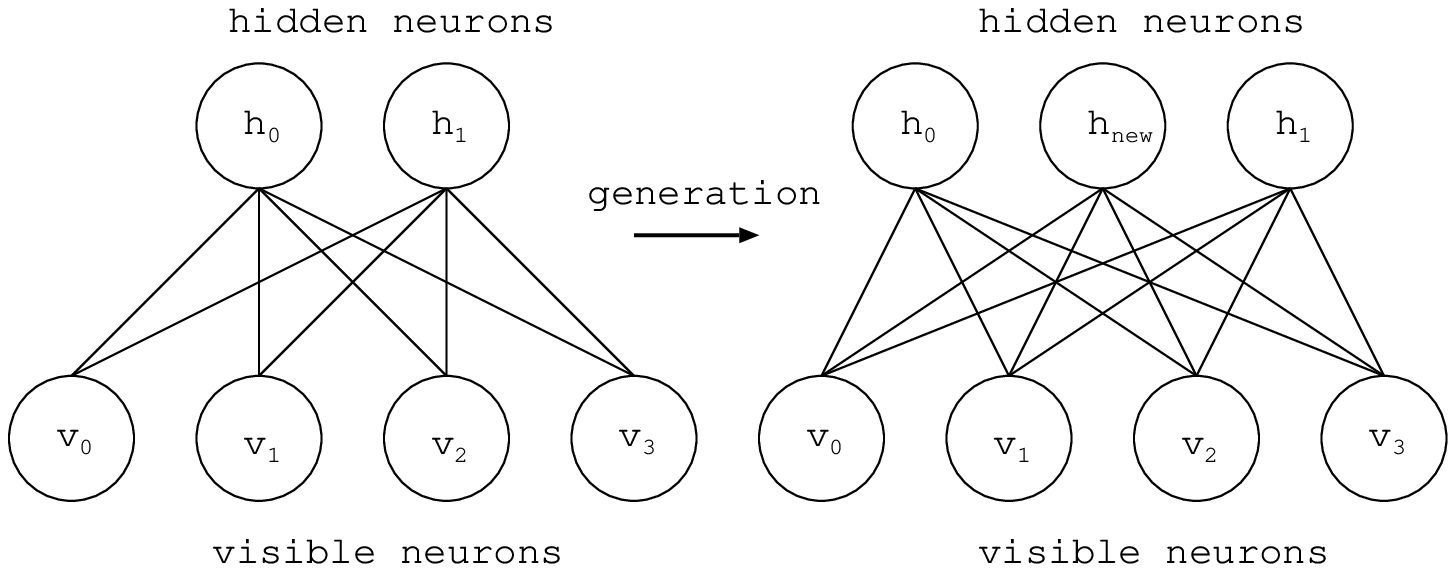}
  \label{fig:neuron_generation}
}
\subfigure[Neuron Annihilation]{\includegraphics[scale=0.5]{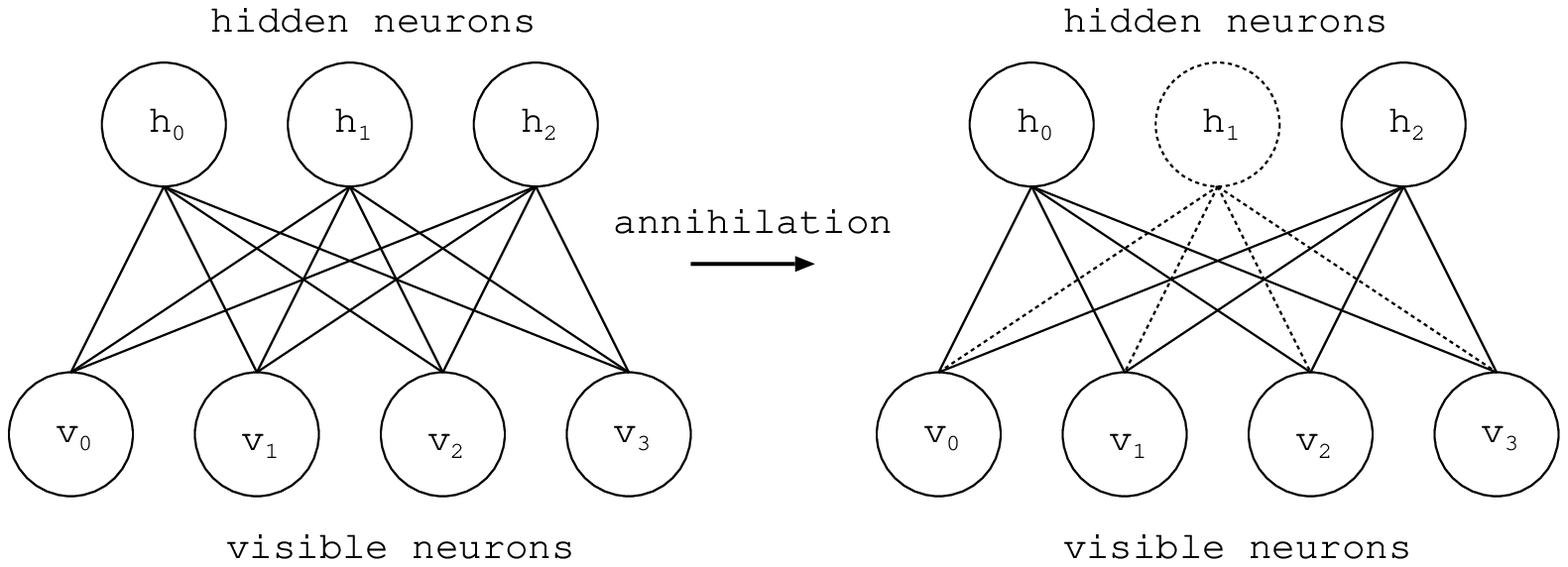}
  \label{fig:neuron_annihilation}
}
\vspace{-3mm}
\caption{Adaptive RBM}
\label{fig:adaptive_rbm}
\vspace{-3mm}
\end{center}
\end{figure}

We have already proposed the adaptive structure learning method of RBM (Adaptive RBM) by self-organizing the network structure for a given input data \cite{Kamada16_SMC, Kamada16_ICONIP}. The method improved the problem that the traditional RBM cannot change its network structure of hidden neurons during the training. The neuron generation / annihilation algorithm of Adaptive RBM (Fig.\ref{fig:adaptive_rbm}) can determine the suitable number of hidden neurons by monitoring WD (Walking Distance) that is the variance of the parameters during the training.

\begin{figure*}[htbp]
\begin{center}
\includegraphics[scale=0.7]{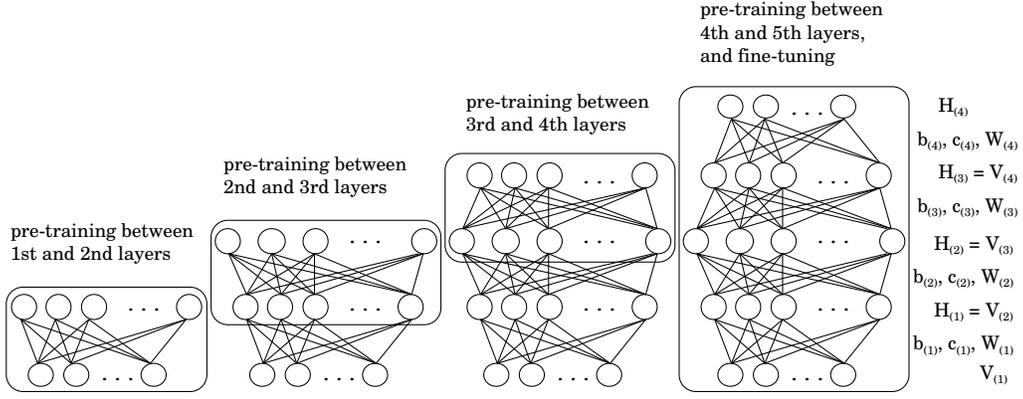}
\caption{An overview of DBN}
\label{fig:dbn}
\end{center}
\end{figure*}

The classification accuracy of a single RBM can be improved by building two or more pre-trained RBMs hierarchically in the DBN architecture \cite{Hinton06} as shown in Fig.\ref{fig:dbn}. We also proposed the adaptive structure learning method of DBN (Adaptive DBN) that can determine the optimal number of hidden layers for a given input data. The developed Adaptive DBN can get obtain higher classification accuracy than the traditional models \cite{Kamada16_TENCON}.

\section{Multi-Modal Data Learning Method}
\label{sec:multi_modal_learning}
In this section, we explain our proposed method that arrange an input sequence based on the relation among the multi-modal data in the training process of Adaptive RBM. First, the structure of multi-modal data which consists of both image data and CSV data such as numerical data, text data, and so on is defined in Section \ref{subsec:multi_modal_learning_data}. Second, the algorithm of the proposed method is described in Section \ref{subsec:multi_modal_learning_algorithm}.

\subsection{Data Structure}
\label{subsec:multi_modal_learning_data}
\begin{figure}[h]
\centering
\includegraphics[scale=0.4]{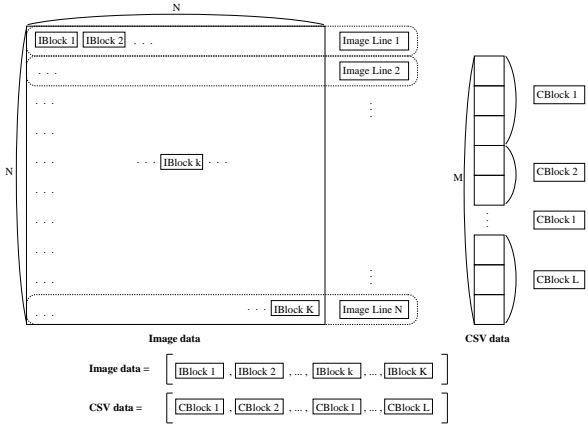}
\vspace{-3mm}
\caption{Data Structure of Image and CSV}
\label{fig:data_structure1}
\end{figure}

Fig.~\ref{fig:data_structure1} shows the data structure of a pair of image data and CSV data. Let an image data be a pixel array with size $N \times N$. Since the visible layer of RBM is represented as one dimensional array, we have to regard two dimensional pixel array as one dimensional array in the training of RBM. In this paper, the sequence of the image data is drawn like a horizontal line from left to bottom right. To be described in Section \ref{subsec:multi_modal_learning_algorithm}, each horizontal line is divided into multiple blocks with the fixed length. Therefore, an image data is defined as $IBlockList =\{ IBlock_{1}, \cdots, IBlock_{k}, \cdots, IBlock_{K}\}$, where $K$ is the number of image blocks. 

Let a CSV data be $L$ numeric items. The value of each item can be converted into a binary pattern with two or more size by cut-off value. Therefore, the CSV data can be represented as a binary vector with size $M$ ($L <= M$) as shown in Fig.~\ref{fig:data_structure1}. By assuming that each numeric item is a CSV block as well as the Image block, the CSV data can be also represented as $CBlockList =\{ CBlock_{1}, \cdots, CBlock_{l}, \cdots, CBlock_{L}\}$.

There may be some implicit relation between an Image block and a CSV block for various kinds of data where 2 or more formats data are recorded separately. Generally, the random sequence of data is given in the learning of big data, but the giving data should be ordered by the implicit relation in sequence. In this paper, the data arrangement method of RBM for an Image block and a CSV block is proposed by following the relation in the training process. RBM trains a binary pattern of input data on some hidden neurons by CD method \cite{Hinton02}, where the difference between a given binary pattern and the represented pattern from the network is trained such as a sensitivity analysis. We consider that the training works for the specific pattern of difference separately. The neuron generation occurs related to such ``sensitive analysis'' of minute change. If the data arrangement is not random sequence, the change itself will depend on the specification of data pattern. Therefore, the data arrangement and neuron generation can find the implicit relation among multi modal data. We will give the mathematical proof in near future. 

\begin{figure}[h]
\centering
\includegraphics[scale=0.7]{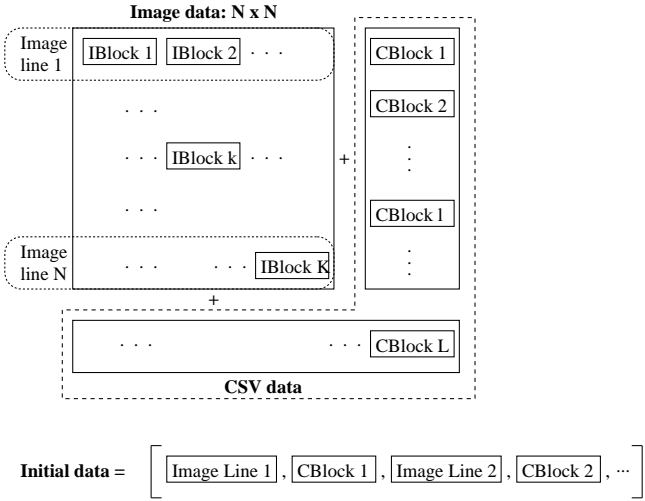}
\vspace{-3mm}
\caption{Initial Arrangement of Image and CSV}
\label{fig:data_structure2}
\end{figure}

Fig.~\ref{fig:data_structure2} shows the initial arrangement of input sequence. To maintain a natural shape of original image data, a line of image data and a block of CSV data are arranged to be select alternately at the initial. Of course, this arrangement is always not the optimal sequence for a given data set. We should consider the method to form the optimal arrangement according to the training situation by sorting them. In Section \ref{subsec:multi_modal_learning_algorithm}, we explain the algorithm of the proposed method that can determine the suitable arrangement during the training. 

\begin{figure}[h]
\centering
\includegraphics[scale=0.45]{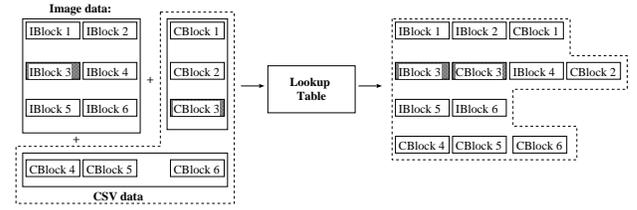}
\vspace{-3mm}
\caption{Mapping with Look up Table}
\label{fig:translation}
\end{figure}

\subsection{Algorithm}
\label{subsec:multi_modal_learning_algorithm}
The basic idea of the sorting process is monitoring which visible neurons are fired as the results of the calculation from hidden neurons to visible neurons followed by CD sampling method. Then, if an image block and a CSV block are fired simultaneously, there will be a relation between their blocks. The neuron fires when the output value of the neuron becomes 1. Because the fired combination can be seen as an occurrence pattern in the input data for the hidden neuron, we can change the sequence of visible neurons that the corresponding visible neurons to be located at more neighborhood position. However, the hidden neuron may have already learned a feature of input during training phase before applying the sorting process. Otherwise, an occurrence pattern acquired from the hidden neuron may not have any meaning relationship. Therefore, our method determines the hidden neuron by monitoring the variance of output value for the hidden neuron. This idea is based on WD of Adaptive RBM.

{\bf Algorithm \ref{alg:rbm}} and {\bf Algorithm \ref{alg:multi-modal}} show the algorithms of our proposed method. The initial arrangement of visible neurons at the beginning is defined as shown in Fig.~\ref{fig:data_structure2}. Please refer our other papers \cite{Kamada16_SMC,Kamada16_TENCON, Ichimura17_IJCNN} for neuron generation / annihilation algorithms of RBM and layer generation condition of DBN for details.

After the sorting process was applied in the algorithm, a table which maps the initial arrangement of input sequence to the converted arrangement is recorded to `Look up Table'. When doing the inference with the trained network, the arrangement of a test data is converted through `Look up Table' as shown in {\bf Algorithm \ref{alg:inference}} and Fig.~\ref{fig:translation}.

\begin{algorithm}                      
\caption{RBM Learning procedure}         
\label{alg:rbm}                          
\begin{algorithmic}[1]
\REQUIRE $\bvec{v} \in \{0, 1 \}^{I}$: Visible layer, $\bvec{h} \in \{0, 1 \}^{J}$: Hidden layer, $\bvec{\theta} \in \{ \bvec{b}, \bvec{c}, \bvec{W}\}$: Learning parameters,\\
  $IBlockList =\{ IBlock_{1}, \cdots, IBlock_{k}, \cdots, IBlock_{K}\}$: Sequence of image blocks, $IBlock_{k}$: Image block, \\
  $CBlockList =\{ CBlock_{1}, \cdots, CBlock_{l}, \cdots, CBlock_{L}\}$: Sequence of csv blocks, $CBlock_{l}$: CSV block. \\
$LookupTable$: Look up table which maps an original input arrangement to the sorted one.
\STATE Initialize the the learning parameters $\bvec{\theta}$.
\STATE Initialize the arrangement of input data and record it to $LookupTable$.
\WHILE{not (termination condition is satisfied)}
\STATE Update the learning parameters $\bvec{\theta}$ for given input data by CD method.
\STATE Calculate the WD of the learning parameters $\bvec{\theta}$ and $\bvec{h}$.
\STATE Execute neuron generation / annihilation algorithms.
\STATE Execute multi-modal data learning method (Algorithm \ref{alg:multi-modal}).
\ENDWHILE
\end{algorithmic}
\end{algorithm}

\begin{algorithm}                      
\caption{Multi Modal Data Learning procedure}         
\label{alg:multi-modal}
\begin{algorithmic}[1]
\STATE Set $\bvec{h}^{S} \in \bvec{h}$ as a set of hidden neurons $h_j$, where $h_j$ equals $1$ and $WD$ for $h_j$ is smaller than the pre-determined value .
\FORALL{$h_{j} \in \bvec{h}^{S}$}
\STATE Calculate a binary pattern of visible layer $\bvec{v}$ from $h_j$.
\STATE Set $IBlockCand \in IBlockList $ and $CBlockCand \in CBlockList $ as a set of candidate Image and CSV blocks for sorting, where $IBlock_{k} = \{v_{i} | v_{i} = 1 \}$ and $CBlock_{l} = \{v_{i} | v_{i} = 1 \}$ are the element of $IBlockCand$ and $CBlockCand$, respectively.
\FORALL{$IBlock_{k} \in IBlockCand$}
\STATE Calculate the neighborhood of $IBlock_{k}$.
\STATE Select $CBlock_l$ from $CBlockCand$ as a CSV block.
\IF{$CBlock_{l}$ is not in the neighborhood of $IBlock_{k}$}
\STATE Change the position of $CBlock_{l}$ to the next to $IBlock_{k}$. 
\STATE Exclude $CBlock_{l}$ from $CBlockCand$ due to the processed block.
\STATE Update $LookupTable$ according to the current arrangement.
\ENDIF
\ENDFOR
\ENDFOR
\end{algorithmic}
\end{algorithm}

\begin{algorithm}                      
\caption{Inference procedure}
\label{alg:inference}                          
\begin{algorithmic}[1]
\REQUIRE $\bvec{V}$: Input Data, $Model$: Trained Network,\\
$LookupTable$: Look up Table which realizes the mapping from original position to the position.  
\STATE Map the given input $\bvec{V}$ to $\bvec{V}^{'}$ through $LookupTable$.
\STATE Execute the inference by giving $\bvec{V}^{'}$ to the trained network $Model$ .
\end{algorithmic}
\end{algorithm}

\section{Experimental Results}
\subsection{Data Sets}
In order to evaluate our proposed method, 3 kinds of data sets, `Medical Data', `CIFAR-10', and `CIFAR-100' were used in the experiment. `Medical Data' is a data set for health check which is provided from `Hiroshima Environment and Health Association \cite{Kanhokyo}'. About 100,000 records were collected from April 2012 to March 2016. Each record consists of 55 test items and medical image data such as chest X-ray. `CIFAR-10' and `CIFAR-100' \cite{CIFAR10} are major image benchmark data sets for classification task. They have 50,000 training records and 10,000 test records of $32 \times 32$ color image. Each image is categorized into one of 10 classes on CIFAR-10, and 100 classes on CIFAR-100. For all the data sets, we made a 10-fold cross validation data set from these data and evaluated them.

For these data sets, the classification accuracy and computational speed (CPU time: sec) were calculated with 3 kinds of methods, `Traditional DBN', `Adaptive DBN', and `Adaptive DBN with Multi-Modal Learning'. For `Medical Data' in the Multi-Modal Learning, the data arrangement of input sequence is arranged based on the relation between an Image block and a CSV block. For `CIFAR-10' and `CIFAR-100', it is arranged based on the relation among Image blocks. The initial arrangement of input sequence is followed in Fig.~\ref{fig:data_structure2}. The following parameters were used for training. The training algorithm is Stochastic Gradient Descent (SGD) method, the batch size is 100, the learning rate is 0.01, and the initial number of hidden neurons is 300. The following GPU workstation was used for training and test. CPU: Intel(R) 24 Core Xeon E5-2670 v3 2.3GHz, GPU: Tesla K80 4992 24GB $\times$ 3, Memory: 64GB, and OS: Cent OS 6.7 64 bit. In addition, the computational speed was also evaluated on a low-end machine to show how our proposed method cuts the time on it. The specification of this machine is as follows. CPU: Intel(R) Core(TM) i5-4460 @ 3.20GHz, GPU: GTX 1080 8GB, Memory: 8GB, and OS: Fedora 23 64 bit.

\subsection{Experimental Results}
Table \ref{tab:accuracy-medical}, Table \ref{tab:accuracy-cifar10}, and Table \ref{tab:accuracy-cifar100} show the experimental results on `Medical Data', `CIFAR-10', and `CIFAR-100'. Each table shows the value of average, standard deviation, maximum, and minimum for test classification accuracy on 10 trials cross validation. `Iterations' is the number of repetitions for training which means the model becomes to satisfy the terminal condition at the value. `Time (Tesla K80)' and `Time (GTX 1080)' show the CPU Time (sec) for training on the two kinds of computers, one is `Tesla K80' and the other is `GTX 1080', respectively.

In the adaptive DBN, 6 layers were automatically generated for the given data set. The classification accuracy (Ave.) at final layer were 0.944, 0.974, and 0.812 for `Medical Data', `CIFAR-10', and `CIFAR-100', respectively. Adaptive DBN obtained higher classification accuracy than Traditional DBN. On the other hand, there was not much difference between Adaptive DBN and the proposed method.

The number of sorting processes in the proposed method was decreased as the layer is generated. Especially, the process was not applied at more 4th layer. We consider that the suitable arrangement for the given data set were almost determined at lower layer. As a result, the proposed method at the higher layer was able to get the fastest computational speed in all the methods even though the lower layer took a much time due to sorting process. To sum up with the total computational speed, the proposed method was able to reduce the speed to 29.4\%, 27.1\%, and 28.0\% for `Medical Data', `CIFAR-10', and `CIFAR-100' compared with Traditional DBN. To translate them to CPU time, they were 67.5(m), 42.0(m), and 42.8(m) for `Tesla K80', 223.1(m), 139.8(m), and 139.3(m) for `GTX 1080'.

\begin{table*}[h]
\caption{Classification Accuracy (Medical Data)}
\vspace{-3mm}
\label{tab:accuracy-medical}
\begin{center}
\scalebox{0.8}[0.8]{
\begin{tabular}{l|r|r|r|r|r|r|r|r|r}
\hline \hline
Model &Layer&Ave.&Std.&Max.&Min.&No. sorting process&Iterations&Time(Tesla K80)&Time(GTX1080) \\ \hline
Traditional DBN&1&0.833&0.011&0.851&0.818&0&500&44.1&151.8 \\ \cline{2-10}
&2&0.862&0.008&0.873&0.850&0&500&37.2&128.3 \\ \cline{2-10}
&3&0.864&0.009&0.877&0.850&0&430&34.7&122.1 \\ \cline{2-10}
&4&0.892&0.010&0.904&0.875&0&456&37.8&130.3 \\ \cline{2-10}
&5&0.907&0.006&0.916&0.893&0&421&36.8&128.9 \\ \cline{2-10}
&6&{\bf 0.911}&0.010&0.930&0.895&0&433&39.1&131.8 \\ \hline
\multicolumn{7}{r}{} & Total &{\bf 229.7}&{\bf 793.0} \\ \hline \hline
Adaptive DBN&1&0.835&0.007&0.847&0.825&0&500&43.1&150.4  \\ \cline{2-10}
&2&0.861&0.008&0.870&0.842&0&440&29.3&99.2 \\ \cline{2-10}
&3&0.864&0.010&0.874&0.845&0&402&31.1&110.0 \\ \cline{2-10}
&4&0.896&0.007&0.907&0.883&0&411&33.5&111.9 \\ \cline{2-10}
&5&0.912&0.012&0.941&0.901&0&422&35.9&120.3 \\ \cline{2-10}
&6&{\bf 0.944}&0.008&0.953&0.932&0&406&35.2&124.0 \\ \hline
\multicolumn{7}{r}{} & Total &{\bf 208.1} &{\bf 715.69} \\ \hline \hline
Multi-Modal Learning&1&0.854&0.012&0.870&0.832&502&413&36.6&130.4  \\ \cline{2-10}
&2&0.879&0.011&0.898&0.861&127&367&25.1&89.2 \\ \cline{2-10}
&3&0.878&0.006&0.888&0.871&23&305&23.7&88.0 \\ \cline{2-10}
&4&0.927&0.009&0.941&0.915&0&299&24.5&86.9 \\ \cline{2-10}
&5&{\bf 0.942}&0.008&0.953&0.925&0&307&25.7&86.3 \\ \cline{2-10}
&6&{\bf 0.942}&0.008&0.953&0.931&0&295&26.6&89.0 \\ \hline
\multicolumn{7}{r}{} & Total &{\bf 162.2}& {\bf 569.8}  \\ \hline
\hline
\end{tabular}
} 
\end{center}
\end{table*}

\begin{table*}[h]
\caption{Classification Accuracy (CIFAR-10)}
\vspace{-3mm}
\label{tab:accuracy-cifar10}
\begin{center}
\scalebox{0.8}[0.8]{
\begin{tabular}{l|r|r|r|r|r|r|r|r|r}
\hline \hline
Model&Layer&Ave.&Std.&Max.&Min.&No. sorting process&Iterations&Time(Tesla K80)&Time(GTX1080) \\ \hline
Traditional DBN&1&0.797&0.011&0.814&0.782&0&500&33.1&116.0\\ \cline{2-10} 
&2&0.808&0.008&0.817&0.788&0&418&22.9&80.9\\ \cline{2-10} 
&3&0.851&0.009&0.866&0.840&0&401&23.7&84.1\\ \cline{2-10} 
&4&0.880&0.015&0.897&0.847&0&378&24.4&87.7\\ \cline{2-10} 
&5&{\bf 0.903}&0.007&0.910&0.888&0&355&25.1&86.2\\ \cline{2-10} 
&6&0.880&0.010&0.894&0.863&0&332&25.9&90.0\\ \hline
\multicolumn{7}{r}{} & Total &{\bf 155.1}&{\bf 545.1}\\ \hline \hline
Adaptive DBN&1&0.817&0.008&0.831&0.804&0&420&30.7&105.8\\ \cline{2-10} 
&2&0.872&0.011&0.888&0.858&0&341&20.4&65.5\\ \cline{2-10} 
&3&0.901&0.008&0.913&0.888&0&367&22.6&80.2\\ \cline{2-10} 
&4&0.952&0.012&0.972&0.937&0&334&22.1&78.9\\ \cline{2-10} 
&5&0.960&0.008&0.978&0.947&0&350&23.4&81.3\\ \cline{2-10} 
&6&{\bf 0.974}&0.010&0.985&0.954&0&326&22.9&77.8\\ \hline
\multicolumn{7}{r}{} & Total &{\bf 142.1}&{\bf 489.5}\\ \hline \hline
Multi-Modal Learning&1&0.833&0.017&0.866&0.817&338&357&26.5&93.3\\ \cline{2-10} 
&2&0.900&0.016&0.921&0.872&31&286&17.1&66.0\\ \cline{2-10} 
&3&0.926&0.013&0.945&0.909&9&304&17.9&62.3\\ \cline{2-10} 
&4&0.965&0.012&0.982&0.947&0&258&16.2&57.7\\ \cline{2-10} 
&5&0.971&0.009&0.984&0.956&0&256&17.5&65.8\\ \cline{2-10} 
&6&{\bf 0.974}&0.010&0.988&0.961&0&244&17.9&60.2\\ \hline
\multicolumn{7}{r}{} & Total &{\bf 113.1}&{\bf 405.3}\\ \hline
\hline
\end{tabular}
} 
\end{center}
\end{table*}

\begin{table*}[h]
\caption{Classification Accuracy (CIFAR-100)}
\vspace{-3mm}
\label{tab:accuracy-cifar100}
\begin{center}
\scalebox{0.8}[0.8]{
\begin{tabular}{l|r|r|r|r|r|r|r|r|r}
\hline \hline
Model&Layer&Ave.&Std.&Max.&Min.&No. sorting process&Iterations&Time(Tesla K80)&Time(GTX1080) \\ \hline
Traditional DBN&1&0.645&0.007&0.667&0.646&0&489&32.7&119.243\\ \cline{2-10} 
&2&0.676&0.015&0.701&0.647&0&411&23.6&76.077\\ \cline{2-10} 
&3&0.701&0.006&0.713&0.695&0&430&24.5&86.213\\ \cline{2-10} 
&4&0.743&0.006&0.749&0.731&0&379&25.7&91.457\\ \cline{2-10} 
&5&{\bf 0.751}&0.011&0.768&0.739&0&361&25.5&85.839\\ \cline{2-10} 
&6&0.748&0.012&0.774&0.733&0&354&26.2&90.039\\ \hline
\multicolumn{7}{r}{} & Total &{\bf 158.2}&{\bf 548.9}\\ \hline \hline
Adaptive DBN&1&0.709&0.009&0.726&0.694&0&429&31.0&106.0\\ \cline{2-10} 
&2&0.740&0.007&0.750&0.727&0&356&21.3&73.9\\ \cline{2-10} 
&3&0.764&0.009&0.780&0.752&0&329&20.9&69.2\\ \cline{2-10} 
&4&0.798&0.013&0.820&0.774&0&338&21.4&76.0\\ \cline{2-10} 
&5&0.807&0.012&0.826&0.790&0&342&22.5&81.0\\ \cline{2-10} 
&6&{\bf 0.812}&0.012&0.834&0.791&0&331&23.5&84.7\\ \hline
\multicolumn{7}{r}{} & Total &{\bf 140.6}&{\bf 490.8}\\ \hline \hline
Multi-Modal Learning&1&0.708&0.007&0.716&0.697&363&370&27.5&98.7\\ \cline{2-10} 
&2&0.785&0.009&0.801&0.773&28&299&17.1&63.8\\ \cline{2-10} 
&3&0.800&0.012&0.821&0.785&4&278&17.3&64.0\\ \cline{2-10} 
&4&0.809&0.007&0.821&0.799&0&256&17.2&55.9\\ \cline{2-10} 
&5&0.815&0.009&0.828&0.801&0&272&18.1&62.0\\ \cline{2-10} 
&6&{\bf 0.823}&0.008&0.834&0.811&0&260&18.2&65.2\\ \hline
\multicolumn{7}{r}{} & Total &{\bf 115.4}&{\bf 409.5} \\ \hline
\hline
\end{tabular}
} 
\end{center}
\end{table*}

\section{Conclusion}
Recently, the techniques of artificial intelligence such as Deep Learning is attracting a lot of attention in not only the research field of computer science but also industrial worlds for applications. Moreover, various kinds of data sets are collected through the Internet with the idea of data collection from IoT devices. In this sense, we have to tackle with the issue of the processing of multi-modal data, which consists of different type of data such as image data, numeric data, and so on. In this paper, the data arrangement are modified according to the similarity of input-output pattern in Adaptive DBN. The similarity of output signals of hidden neurons is made by the order rearrangement of hidden neurons. The experimental results showed that the proposed method realized faster computational speed without decreasing the classification accuracy than the traditional model. In future, the proposed sorting method will be applied to our proposed recurrent model for time-series data set.

\section*{Acknowledgment}
This work was supported by JAPAN MIC SCOPE Grand Number 162308002, Artificial Intelligence Research Promotion Foundation, and JSPS KAKENHI Grant Number JP17J11178.

\end{document}